\documentclass[10pt,twocolumn,letterpaper]{article}

\usepackage{iccv}              

\definecolor{iccvblue}{rgb}{0.21,0.49,0.74}
\usepackage[pagebackref,breaklinks,colorlinks,allcolors=iccvblue]{hyperref}
\usepackage{graphicx}
\usepackage{animate}
\usepackage{tabularx}
\usepackage{pifont}

\newcommand{\tablestyle}[2]{\setlength{\tabcolsep}{#1}\renewcommand{\arraystretch}{#2}\centering\footnotesize}
\newlength\savewidth

\newcommand{\animationNotes}{
\textbf{\textit{Please open in Acrobat Reader and click the image to play the animation.}}
}
\newcommand{\numColumns}{4}
\newcommand{\columnSpacing}{0.5em}

\usepackage{multirow}
\def\paperID{3017} 
\def\confName{ICCV}
\def\confYear{2025}

\title{Can video generation replace cinematographers? Research on the cinematic
language of generated video}

\author{
Xiaozhe Li\textsuperscript{\rm 1,}\thanks{Equal contribution.}, 
Kai Wu\textsuperscript{\rm 2,}\footnotemark[1],
Siyi Yang\textsuperscript{\rm 1},
Yizhan Qu\textsuperscript{\rm 1},
Guohua Zhang\textsuperscript{\rm 1},
Zhiyu Chen\textsuperscript{\rm 1},\\
Jiayao Li\textsuperscript{\rm 1},
Jiangchuan Mu\textsuperscript{\rm 1},
Xiaobin Hu\textsuperscript{\rm 3},
Wen Fang\textsuperscript{\rm 1},\\
Mingliang Xiong\textsuperscript{\rm 1},
Hao Deng\textsuperscript{\rm 1,}\thanks{Corresponding authors.},
Qingwen Liu\textsuperscript{\rm 1,}\footnotemark[2],
Gang Li\textsuperscript{\rm 1},
Bin He\textsuperscript{\rm 1}\\
\textsuperscript{\rm 1}Tongji University \quad \
\textsuperscript{\rm 2}ByteDance \quad \
\textsuperscript{\rm 3}Technical University of Munich \\
{\tt\small \{Lxxzzz,2253110,2251645, 2151368, 2354271, 2351405, 2411941\}@tongji.edu.cn} \\
{\tt\small \{wen.fang,denghao1984,qliu, lig, hebin\}@tongji.edu.cn} \\  
{\tt\small wukaiwork@outlook.com} \quad
{\tt\small xiaobin.hu@tum.de}\quad
{\tt\small xiongml@foxmail.com}
}

\begin{document}

\twocolumn[{%
\renewcommand\twocolumn[1][]{#1}%
\maketitle
\vspace{-3em}
\begin{center}
    \centering
    \captionsetup{type=figure}
    \includegraphics[width=0.87\linewidth]{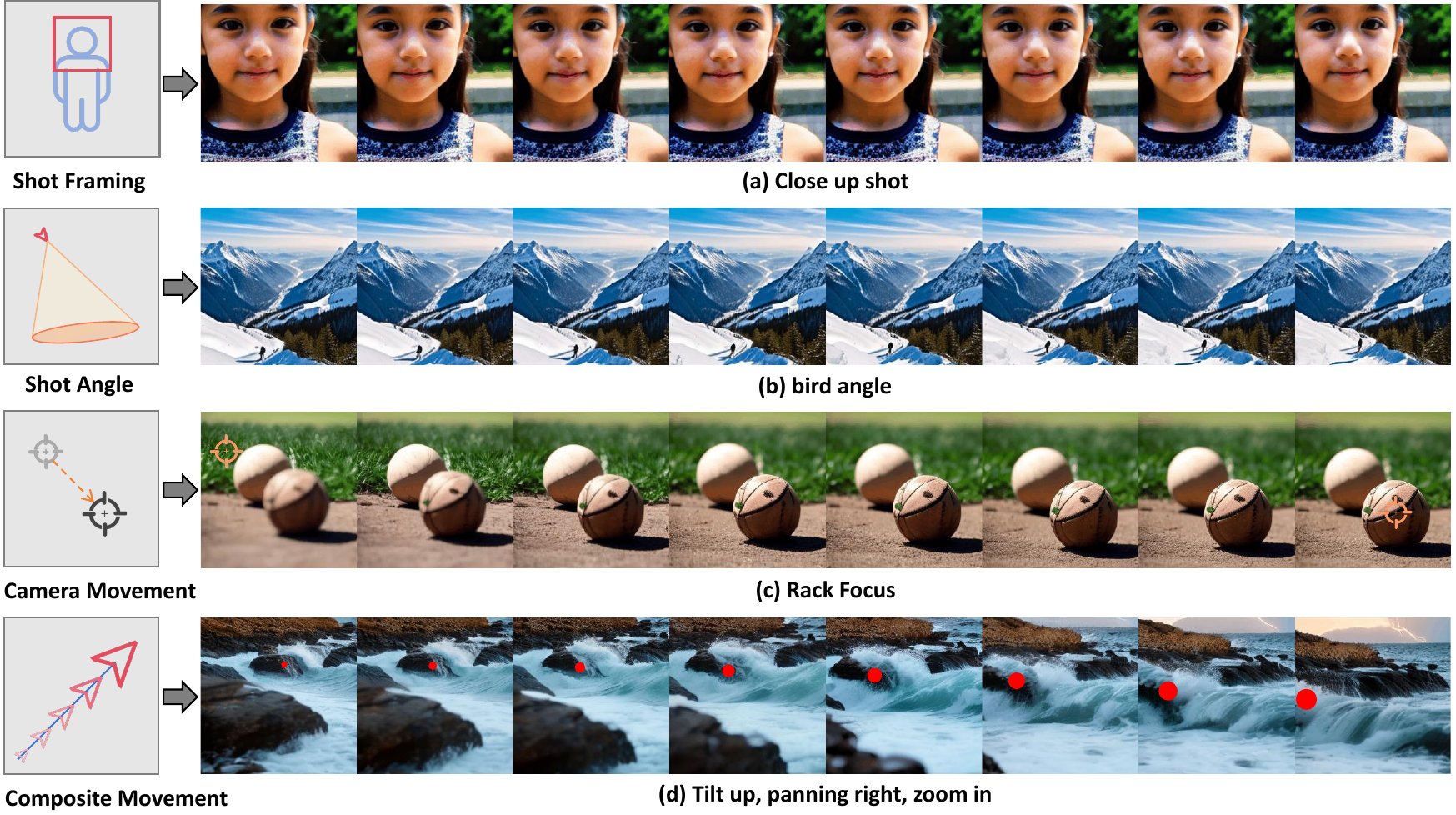}
    \vspace{-6pt}
    \caption{\textbf{Cinematic language generation results of CameraDiff.} CameraDiff generates twenty types of cinematic controls, including shot framing (a), shot angles (b), and camera movements (c) such as lens shifts (e.g., rack focus, zoom in/out) and camera body movements (e.g., tilt up/down). Additionally, it enables the flexible combination of multiple camera movements (d).}
    \label{fig:spotlight}

\end{center}%
}
]

\begin{abstract}

Recent advancements in text-to-video (T2V) generation have leveraged diffusion models to enhance visual coherence in videos synthesized from textual descriptions. However, existing research primarily focuses on object motion, often overlooking cinematic language, which is crucial for conveying emotion and narrative pacing in cinematography. To address this, we propose a threefold approach to improve cinematic control in T2V models.
First, we introduce a meticulously annotated cinematic language dataset with twenty subcategories, covering shot framing, shot angles, and camera movements, enabling models to learn diverse cinematic styles. Second, we present CameraDiff, which employs LoRA for precise and stable cinematic control, ensuring flexible shot generation. Third, we propose CameraCLIP, designed to evaluate cinematic alignment and guide multi-shot composition.
Building on CameraCLIP, we introduce CLIPLoRA, a CLIP-guided dynamic LoRA composition method that adaptively fuses multiple pre-trained cinematic LoRAs, enabling smooth transitions and seamless style blending. Experimental results demonstrate that CameraDiff ensures stable and precise cinematic control, CameraCLIP achieves an R@1 score of 0.83, and CLIPLoRA significantly enhances multi-shot composition within a single video, bridging the gap between automated video generation and professional cinematography.\textsuperscript{1}
\footnotetext[1]{Codes, data, and model weights will be open-sourced after the paper is accepted.}
\end{abstract}
\section{Introduction}
\begin{figure*}[t!]
    \centering
    \includegraphics[width=\textwidth]{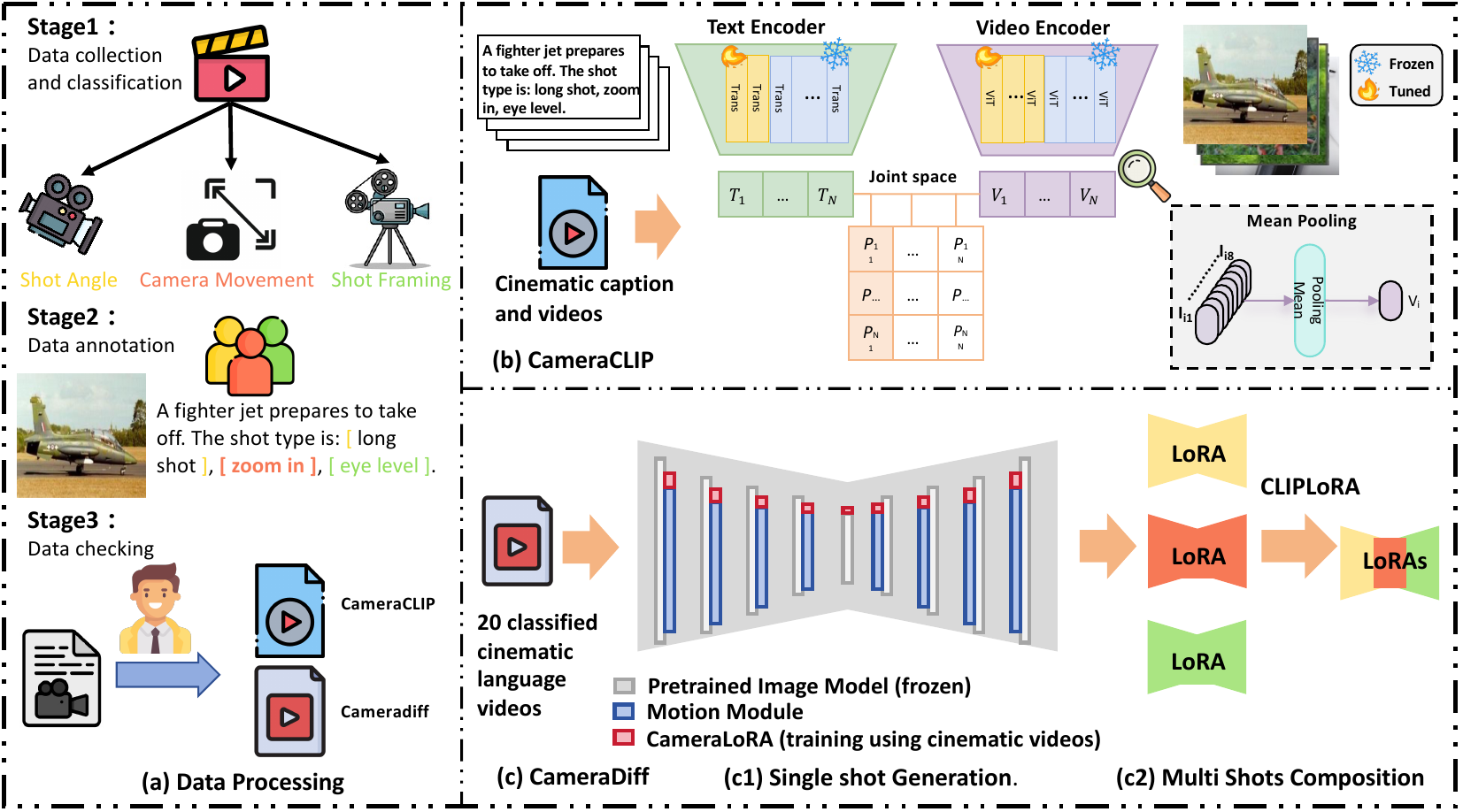}
    \caption{\textbf{Pipeline of our threefold approach.} (a) Data processing: Stage 1—data collection and classification, Stage 2—human annotation, and Stage 3—manual verification. (b) CameraCLIP training: The text encoder is trained on the last two layers, while the image encoder is trained on the last four layers. Each video is uniformly sampled into eight frames, encoded via the image encoder, and mean-pooled to obtain video features. These features, combined with text features, undergo contrastive learning in a joint space to enhance similarity. (c) CameraDiff: LoRA enables single-shot cinematic control, while CLIPLoRA facilitates multi-shot composition within a single video.}
    \label{fig:pipeline}
    \vspace{-2em}
\end{figure*}

Recent advancements in text-to-video (T2V) generation have leveraged diffusion models to enhance the visual coherence of videos synthesized from textual descriptions~\cite{ho2020denoising, rombach2022stable_diffussion, saharia2022photorealistic, blattmann2023align, chen2023videocrafter1, ho2022imagen, wang2023videocomposer}. However, most T2V research has primarily focused on object motion, prioritizing content alignment and visual consistency. Additionally, the high labor costs associated with collecting cinematic language data from real-world scenarios have hindered progress in text-controlled cinematic video generation, leaving it largely unexplored.

Current T2V models cannot replace skilled cinematographers due to their limited ability to generate complex cinematic language in videos.
Previous works, such as AnimateDiff~\cite{guo2023animatediff}, introduced motion priors to enhance animation fluidity and utilized MotionLoRA for efficient fine-tuning of motion patterns. However, these methods restricted camera motion to basic operations like zooming and rotation. MotionCtrl~\cite{wang2024motionctrl} introduced modules for independent control of camera and object motion, while Direct-a-Video~\cite{Yang_2024Direct-a-Video} aimed to replicate real-world video shooting by decoupling object and camera motion, enabling user control over translation, zoom, and directional movement. Despite these advancements, both MotionCtrl and Direct-a-Video struggle to generate a full range of cinematic styles, particularly in shot framing and shot angle control, and require additional camera parameter inputs to control camera movement, limiting flexible camera manipulation.

While these advancements have established a foundation for camera control in T2V models, significant limitations persist in generating diverse cinematic styles. The key challenges are: (1) the lack of a high-quality cinematic language dataset and (2) the absence of a method for multi-cinematic language composition to flexibly generate multi-shot videos. For example, producing a "long shot, eye-level, panning right, zoom-in, and tilt-up view of a bear walking in the snow near a lake" remains difficult. To address these challenges, we propose a threefold approach to enhance T2V models' capability to generate high-quality, multi-shot composition videos.

Firstly, we introduce Cinematic2K, a comprehensive video cinematic language dataset comprising twenty subcategories and approximately 2,000 videos covering shot framing, shot angles, and camera movements. The dataset consists of video-text pairs representing diverse shot types captured from real-world scenarios. To ensure high quality, we perform meticulous human annotation and verification.

Secondly, we propose CameraDiff, which leverages Cinematic2K to enable precise control over cinematic language patterns, encompassing twenty cinematic types. This includes four shot framing styles, five shot angles, and eleven camera movements, such as lens shifts and camera body movements.

Finally, we introduce CameraCLIP, which enhances the CLIP~\cite{Radford2021clip_icml} model’s ability to comprehend complex cinematic language in video content. Building on CameraCLIP, we propose CLIPLoRA, which employs CameraCLIP as an evaluator to dynamically guide LoRA composition. This approach adaptively selects the optimal pre-trained cinematic LoRAs for activation during the denoising stage, ensuring smooth transitions and seamless blending of cinematic styles within a single video.

Our specific contributions can be summarized as follows:

    
\begin{itemize}
    \item We introduce Cinematic2K, a cinematic language video dataset comprising approximately 2,000 high-quality videos across twenty cinematic subcategories, including shot framing, shot angles, and camera movements. Additionally, we propose CameraDiff, which enables the generation of twenty distinct cinematic patterns.
    
    \item We introduce CameraCLIP, a model with a strong capability to align complex cinematic language with corresponding videos. Building on this, we propose CLIPLoRA, a CLIP-guided LoRA composition method that enables seamless multi-shot fusion within a single video.
    
    \item Experimental results show that CameraCLIP outperforms popular models in aligning videos with their cinematic language descriptions, achieving an R@1 score of 0.83. CameraDiff provides stable control over twenty shot types, while CLIPLoRA effectively composites multiple LoRAs. 
\end{itemize}
\section{Related Works}

\textbf{Image/Video CLIP Models.}
CLIP~\cite{Radford2021clip_icml} is a multi-modal model based on contrastive learning, commonly used to assess image-text similarity. Trained on large-scale text-image pairs, CLIP excels in zero-shot generalization, enabling its application to tasks such as detection~\cite{2021guopeniclr,li2022groundedcvpr}, segmentation~\cite{xu2022groupvit, lilanguage}, video understanding~\cite{luo2021clip4clip,xu2021videoclip}, retrieval~\cite{wang2023cocaclip}, and image generation~\cite{ramesh2022hierarchicaltext, frans2022clipdraw,crowson2022vqgan, vinker2022clipasso}.For video analysis, ViCLIP~\cite{wang2023internvid} enhances video analysis by incorporating spatio-temporal attention and partial random patch masking, while Long-CLIP~\cite{zhang2024longclip} addresses embedding length limitations. VideoCLIP-XL~\cite{wang2024videoclipxl} improves alignment for long-form video descriptions via text-similarity-guided matching. 
Despite these advancements, existing models focus on overall video semantics or long-text inputs, lacking specific attention to cinematic language like shot framing, angle, and camera movement. To fill this gap, we introduce CameraCLIP, which enhances the model's understanding of these cinematic elements.

\noindent\textbf{Text-to-Video Generation for Camera Control.}
Early video generation research relied on GANs and VAEs~\cite{gnn2017iccv, skorokhodov2022stylegan, tulyakov2018mocogan, vondrick2016generating}. With the success of diffusion models in high-fidelity image generation~\cite{ho2020denoising, rombach2022stable_diffussion, saharia2022photorealistic}, diffusion-based approaches have gained traction for video generation, improving visual coherence in text or image-guided content~\cite{blattmann2023align, chen2023videocrafter1, ho2022imagen, wang2023videocomposer, blattmann2023stable}. 
Recent works in T2V models have focused on camera control. AnimateDiff~\cite{guo2023animatediff} introduced motion priors for smoother animations, while CameraCTRL~\cite{he2024cameractrl}, MotionCtrl~\cite{wang2024motionctrl} and Direct-a-Video~\cite{Yang_2024Direct-a-Video} offered more independent control over camera and object motion. However, AnimateDiff is limited to basic camera movements, and CameraCTRL, MotionCtrl and Direct-a-Video require extra camera parameters, restricting flexible control. 

\noindent\textbf{Cinematic Dataset for T2V Model.}
Despite the availability of high-quality datasets for image/video understanding~\cite{chen2011collecting, chen2024sharegpt4video, nan2024openvid1mlargescalehighqualitydataset, li2024llavaonevision} and generation~\cite{schuhmann2022laion, bain2021frozen}, comprehensive datasets for cinematic language understanding and generation remain scarce. RealEstate10K~\cite{zhou2018stereoRealestate10k} captures only camera movement trajectories but includes numerous irregular motions, making precise control over basic operations like zooming or panning challenging. CineScale~\cite{savardi2021cinescale} and CineScale2~\cite{savardi2023cinescale2} focus on shot framing and angles, respectively, yet lack detailed camera movement information, as shown in Table~\ref{tab:camera_datasets}, thereby limiting expressive cinematic video generation.


\noindent\textbf{Multi-LoRA Composition.}
Low-Rank Adaptation (LoRA)~\cite{hulora} has shown great potential in text-to-image/video generation, enabling high-quality content with low computational cost. Building on this, LoRA Merge~\cite{ryu2023loramerge} was introduced to combine multiple LoRAs into a unified LoRA for diffusion models. Zhong~\cite{zhong2024multiloracomposition} expanded this with LoRA Switch and LoRA Composite,  which focus on the diffusion denoising stage. LoRA Switch alternates LoRAs at specific denoising steps, while LoRA Composite averages guidance scores to improve composition quality and robustness
However, these methods are static and lack dynamic adaptation to specific tasks, limiting their ability to model complex cinematic language. To address this, we propose CLIPLoRA, a CLIP-guided composition approach that leverages cameraCLIP as an evaluator to optimize LoRA selection during diffusion, enhancing the realism of camera motion in video generation.
\section{Method}
Our threefold approach begins with data collection and processing to construct Cinematic2K (Section~\ref{sec:data}). Using cinematic captions and videos from Cinematic2K, we propose CameraCLIP to enhance cinematic text-video alignment (Section~\ref{sec:cameraclip}). Finally, we introduce CameraDiff, which first enables single-shot control using twenty classified cinematic videos and then utilizes CameraCLIP as an evaluator to guide multi-shot composition (Section~\ref{sec:cameradiff}).
\subsection{Cinematic Language Dataset Construction}
\label{sec:data}
\begin{table}[ht]
    \centering
    \begin{tabular}{lcc}
        \toprule
        Dataset  & M/F/A& Annotation \\
        \midrule
        RealEstate10K \cite{zhou2018stereoRealestate10k}  & \ding{51} , \ding{55} , \ding{55}  & \ding{55} \\
        CineScale \cite{savardi2021cinescale}  & \ding{55} , \ding{51} , \ding{55}  & \ding{51} \\
        CineScale2 \cite{savardi2023cinescale2}  & \ding{55} , \ding{55} , \ding{51}  & \ding{51} \\
        \midrule
        Cinematic2K & \ding{51} , \ding{51} , \ding{51}  & \ding{51} \\
        \bottomrule
    \end{tabular}
    \caption{Comparison of video datasets in terms of cinematic language attributes, including camera movement (M), shot framing (F), and shot angle (A), as well as the availability of annotations. Cinematic2K provides the most comprehensive coverage, supporting all three attributes with full annotation, whereas other datasets lack complete coverage of these aspects.}
    \label{tab:camera_datasets}
    \vspace{-1em}
\end{table}
As shown in Table~\ref{tab:camera_datasets}, existing cinematic datasets are not systematically organized. These datasets either do not contain sufficient cinematic language or lack specific cinematic types with their own annotations. Therefore, it is necessary to construct the cinematic language dataset to bridge this gap. We sourced videos from Pexels\textsuperscript{1} and Videvo\textsuperscript{2}, two platforms providing free high-quality video data. The dataset construction process, as illustrated in Figure~\ref{fig:pipeline} (a), was divided into three steps.

\begin{figure}[t]
    \centering
    \includegraphics[width=0.8\columnwidth]{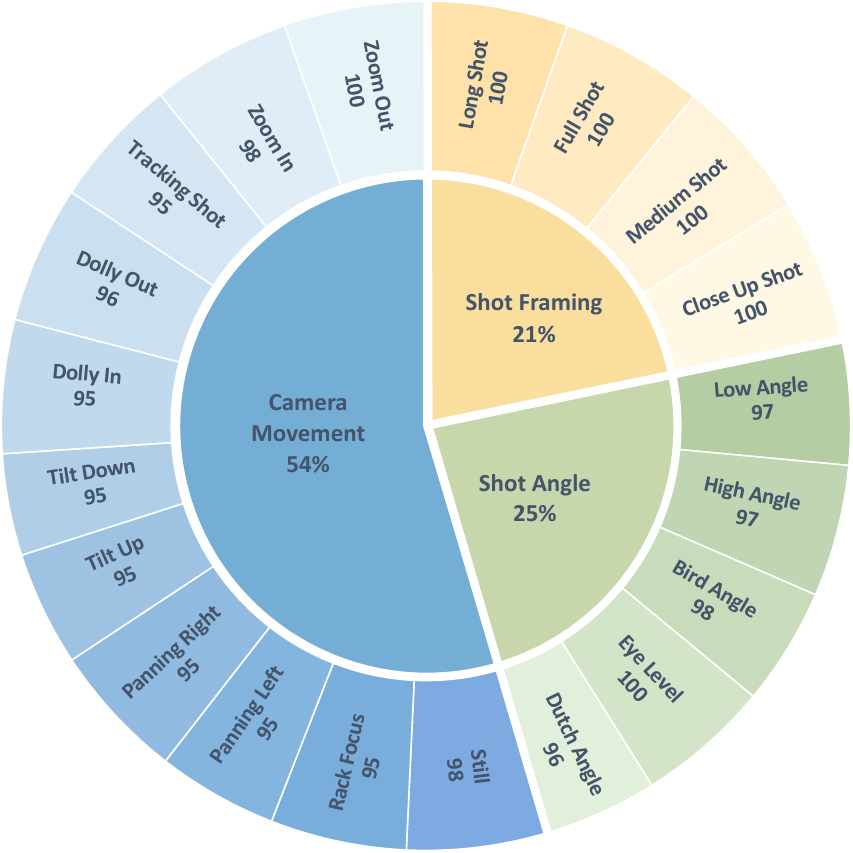}
    \caption{This dataset contains three categories of cinematic language: shot framing, shot angle, and camera movement. It consists of 20 subclasses and contains approximately 2,000 entries, systematically covering all classifications of cinematic language.}
    \label{fig:pie_distribution_one}
    \vspace{-1.5em}
\end{figure}

\footnotetext[1]{https://www.pexels.com}
\footnotetext[2]{https://www.videvo.net}
\noindent\textbf{Step I: Cinematic Language Classification and Video Data Collection.} 
We categorized cinematic language into three primary types: shot framing, shot angle, and camera movement. To efficiently build a high-quality cinematic language dataset on a large scale, we employed automated tools to extract content descriptions and cinematic language annotations for each video. In total, we gathered approximately 2,000 video samples. The detailed data distribution of the cinematic language dataset is shown in Figure~\ref{fig:pie_distribution_one}. \textsuperscript{3}
\footnotetext[3]{Detailed descriptions of these 20 cinematic language types are provided in the Supplementary Material.}

\noindent\textbf{Step II: Fine-grained Annotation of Cinematic Language Intervals.}  
Since the collected data originates from real-world scenarios, it may contain noise, such as inconsistent camera movements or transitions in shot types like 'rack focus,' which typically unfold over a few seconds. To enhance data quality, we manually annotated each video after Step I, precisely marking the start and end intervals of these fine-grained camera movement transitions.

\noindent\textbf{Step III: Verification of Cinematic Language Annotations.}  
As each video's cinematic language comprises shot framing, shot angle, and camera movement, automatically generated descriptions from online sources often contain inaccuracies or omit certain shot types. To address this, after completing the annotation process in Step II, we manually proofread and refine the dataset, supplementing missing shot types to ensure accuracy and completeness. Following human annotation, each cinematic entry is structured as: "shot angle" + "camera movement" + "shot framing." Additionally, videos with distinct cinematic features are labeled as "Typical Video," with a complete example provided in the Supplementary Materials.

After the above data processing, we construct Cinematic2K, a dataset of approximately 2,000 high-quality cinematic language videos with meticulous human annotation and verification. Notably, real-world camera movement is complex, and excessive irrelevant motion can negatively impact data quality. Therefore, Cinematic2K strikes a balance between data quality and quantity to ensure reliable annotations while maintaining a diverse dataset.

\begin{figure*}[htbp!]
    \centering
    \renewcommand{\numColumns}{4}
    \renewcommand{\columnSpacing}{0.5em}
    \begin{tabular}{
        @{}
        p{\dimexpr(\textwidth-\columnSpacing*(\numColumns-1))/\numColumns} @{\hspace{\columnSpacing}}
        p{\dimexpr(\textwidth-\columnSpacing*(\numColumns-1))/\numColumns} @{\hspace{\columnSpacing}}
        p{\dimexpr(\textwidth-\columnSpacing*(\numColumns-1))/\numColumns} @{\hspace{\columnSpacing}}
        p{\dimexpr(\textwidth-\columnSpacing*(\numColumns-1))/\numColumns} @{}
    }
        \animategraphics[width=\linewidth]{8}{figs/close_up_shot/frame_}{00}{15} & 
        \animategraphics[width=\linewidth]{8}{figs/medium_shot/frame_}{00}{15} & 
        \animategraphics[width=\linewidth]{8}{figs/full_shot/frame_}{00}{15} & 
        \animategraphics[width=\linewidth]{8}{figs/long_shot/frame_}{00}{15}
    \end{tabular}
    
    \begin{tabularx}{\textwidth}{XXXX}
        \centering \emph{\small \textbf{close up shot}} & 
        \centering \emph{\small \textbf{medium shot}}  &
        \centering \emph{\small \textbf{full shot}}  &
        \centering \emph{\small \textbf{long shot}} 
    \end{tabularx}

    \renewcommand{\numColumns}{5}
    \renewcommand{\columnSpacing}{0.25em}
    \begin{tabular}{
        @{}
        p{\dimexpr(\textwidth-\columnSpacing*(\numColumns-1))/\numColumns} @{\hspace{\columnSpacing}}
        p{\dimexpr(\textwidth-\columnSpacing*(\numColumns-1))/\numColumns} @{\hspace{\columnSpacing}}
        p{\dimexpr(\textwidth-\columnSpacing*(\numColumns-1))/\numColumns} @{\hspace{\columnSpacing}}
        p{\dimexpr(\textwidth-\columnSpacing*(\numColumns-1))/\numColumns} @{\hspace{\columnSpacing}}
        p{\dimexpr(\textwidth-\columnSpacing*(\numColumns-1))/\numColumns} @{}
    }
        \animategraphics[width=\linewidth]{8}{figs/low_angle/frame_}{00}{15} & 
        \animategraphics[width=\linewidth]{8}{figs/eye_shot/frame_}{00}{15} & 
        \animategraphics[width=\linewidth]{8}{figs/high_angle/frame_}{00}{15} & 
        \animategraphics[width=\linewidth]{8}{figs/dutch_angle/frame_}{00}{15} & 
        \animategraphics[width=\linewidth]{8}{figs/bird_angle/frame_}{00}{15}
    \end{tabular}
    
    \begin{tabularx}{\textwidth}{XXXXX}
        \centering \emph{\small \textbf{low angle}}     & 
        \centering \emph{\small \textbf{eye level}}     & 
        \centering \emph{\small \textbf{high angle}}    & 
        \centering \emph{\small \textbf{dutch angle}}   & 
        \centering \emph{\small \textbf{bird angle}}
    \end{tabularx}
    
    \renewcommand{\numColumns}{6}
    \renewcommand{\columnSpacing}{0.25em}
    \begin{tabular}{
        @{}
        p{\dimexpr(\textwidth-\columnSpacing*(\numColumns-1))/\numColumns} @{\hspace{\columnSpacing}}
        p{\dimexpr(\textwidth-\columnSpacing*(\numColumns-1))/\numColumns} @{\hspace{\columnSpacing}}
        p{\dimexpr(\textwidth-\columnSpacing*(\numColumns-1))/\numColumns} @{\hspace{\columnSpacing}}
        p{\dimexpr(\textwidth-\columnSpacing*(\numColumns-1))/\numColumns} @{\hspace{\columnSpacing}}
        p{\dimexpr(\textwidth-\columnSpacing*(\numColumns-1))/\numColumns} @{\hspace{\columnSpacing}}
        p{\dimexpr(\textwidth-\columnSpacing*(\numColumns-1))/\numColumns} @{}
    }
        \animategraphics[width=\linewidth]{8}{figs/dolly_in/frame_}{00}{15} & 
        \animategraphics[width=\linewidth]{8}{figs/dolly_out/frame_}{00}{15} & 
        \animategraphics[width=\linewidth]{8}{figs/panning_left/frame_}{00}{15} & 
        \animategraphics[width=\linewidth]{8}{figs/panning_right/frame_}{00}{15} & 
        \animategraphics[width=\linewidth]{8}{figs/rack_focus/frame_}{00}{15} & 
        \animategraphics[width=\linewidth]{8}{figs/still/frame_}{00}{15}
    \end{tabular}
    
    \begin{tabularx}{\textwidth}{XXXXXX}
        \centering \emph{\small \textbf{dolly in}}     & 
        \centering \emph{\small \textbf{dolly out}}    & 
        \centering \emph{\small \textbf{panning left}} &
        \centering \emph{\small \textbf{panning right}} &
        \centering \emph{\small \textbf{rack focus}}   &
        \centering \emph{\small \textbf{still}}
    \end{tabularx}

    \renewcommand{\numColumns}{5}
    \renewcommand{\columnSpacing}{0.25em}
    \begin{tabular}{
        @{}
        p{\dimexpr(\textwidth-\columnSpacing*(\numColumns-1))/\numColumns} @{\hspace{\columnSpacing}}
        p{\dimexpr(\textwidth-\columnSpacing*(\numColumns-1))/\numColumns} @{\hspace{\columnSpacing}}
        p{\dimexpr(\textwidth-\columnSpacing*(\numColumns-1))/\numColumns} @{\hspace{\columnSpacing}}
        p{\dimexpr(\textwidth-\columnSpacing*(\numColumns-1))/\numColumns} @{\hspace{\columnSpacing}}
        p{\dimexpr(\textwidth-\columnSpacing*(\numColumns-1))/\numColumns} @{}
    }
        \animategraphics[width=\linewidth]{8}{figs/tilt_up/frame_}{00}{15} & 
        \animategraphics[width=\linewidth]{8}{figs/tilt_down/frame_}{00}{15} & 
        \animategraphics[width=\linewidth]{8}{figs/tracking_shot/frame_}{00}{15} & 
        \animategraphics[width=\linewidth]{8}{figs/zoom_in/frame_}{00}{15} & 
        \animategraphics[width=\linewidth]{8}{figs/zoom_out/frame_}{00}{15}
    \end{tabular}
    
    \begin{tabularx}{\textwidth}{XXXXX}
        \centering \emph{\small \textbf{tilt up}}      & 
        \centering \emph{\small \textbf{tilt down}}    & 
        \centering \emph{\small \textbf{tracking shot}} &
        \centering \emph{\small \textbf{zoom in}}     & 
        \centering \emph{\small \textbf{zoom out}}
    \end{tabularx}
    \vspace{-1.5em}
    \caption{
        \textbf{Qualitative results of single-shot generation in CameraDiff.} CameraDiff enables the generation of specific cinematic language for individual shot types. The first and second rows illustrate control over shot framing and shot angles, respectively, while the third and fourth rows demonstrate control over camera movements. Each cinematic type is annotated below the figure. \animationNotes
    }
    \label{fig:single_lora}
    \vspace{-1.5em}
\end{figure*}

\subsection{CameraCLIP}
\label{sec:cameraclip}
\begin{figure*}[htbp]
    \centering
    \renewcommand{\numColumns}{5}
    \renewcommand{\columnSpacing}{0.25em}
    \begin{tabular}{
        @{}
        p{\dimexpr(\textwidth-\columnSpacing*(\numColumns-1))/\numColumns} @{\hspace{\columnSpacing}}
        p{\dimexpr(\textwidth-\columnSpacing*(\numColumns-1))/\numColumns} @{\hspace{\columnSpacing}}
        p{\dimexpr(\textwidth-\columnSpacing*(\numColumns-1))/\numColumns} @{\hspace{\columnSpacing}}
        p{\dimexpr(\textwidth-\columnSpacing*(\numColumns-1))/\numColumns} @{\hspace{\columnSpacing}}
        p{\dimexpr(\textwidth-\columnSpacing*(\numColumns-1))/\numColumns} @{}
    }
        \animategraphics[width=\linewidth]{8}{figs/td_pl/frame_}{00}{15} & 
        \animategraphics[width=\linewidth]{8}{figs/pr_zoomin/frame_}{00}{15}  & 
        \animategraphics[width=\linewidth]{8}{figs/zoomout_pl/frame_}{00}{15} & 
        \animategraphics[width=\linewidth]{8}{figs/zoomin_tilt_up_pr/frame_}{00}{15}& 
        \animategraphics[width=\linewidth]{8}{figs/zoomout_td_pl/frame_}{00}{15}
    \end{tabular}
    
    \begin{tabularx}{\textwidth}{XXXXX}
        \centering \emph{\small \textbf{Long shot, low angle, panning left and tilt down}}     & 
        \centering \emph{\small \textbf{Long shot, high level, zoom in and panning right}}       & 
        \centering \emph{\small \textbf{Long shot, eye level, zoom out and panning left}}   & 
        \centering \emph{\small \textbf{Full shot, eye level, zoom in, panning right and tilt up}}& 
        \centering \emph{\small \textbf{Long shot, low angle, zoom out, panning left and tilt down}} 
    \end{tabularx}
    \vspace{-1em}
    \caption{
        \textbf{Qualitative results of multi-shot composition in CameraDiff.} We combine single-shot LoRAs using CLIPLoRA to achieve blending of multiple shots within a single video, with cinematic language details provided below each figure. \animationNotes
    }
    \vspace{-1.5em}
    \label{fig:multi_lora}
\end{figure*}

Our CameraCLIP extends the capabilities of the pre-trained CLIP model~\cite{Radford2021clip_icml} to the domain of cinematic language video-text alignment. Figure~\ref{fig:pipeline} (b) presents the architecture of our model, with its key components described in detail below.

\noindent\textbf{Text Encoder}.  
To adapt the CLIP model to our task while preserving its strong generalization capabilities, we fine-tune only the last two layers (layers 10 and 11) while freezing all earlier layers. This selective fine-tuning allows the text encoder to effectively capture cinematic language without overfitting, maintaining its broader linguistic representation.

\noindent\textbf{Vision Transformer (ViT) Encoder}.  
To obtain a compact representation of the entire video, we sample 8 frames per video, resize and embed them, and then apply mean pooling to aggregate their features:

\[
V = \frac{1}{N} \sum_{i=1}^{N} I_i,
\]

where \( N \) denotes the number of sampled frames, and \( I_i \) represents the feature of the \( i \)-th frame. This approach balances information coverage and computational efficiency. To further adapt the ViT encoder to cinematic language tasks, we fine-tune its higher layers (layers 20 to 23), refining its visual representations while preserving the pre-trained model's generalization ability.

\noindent\textbf{Text-Video Similarity Computation}.  
The similarity between the video feature vector \( V \) and the text feature vector \( T \) is computed using cosine similarity:

\begin{equation}
    \text{cosine\_sim}(V, T) = \frac{V \cdot T}{\|V\|_2 \|T\|_2},
\end{equation}

where:
\begin{itemize}
    \item \( V \) and \( T \) are the video and text feature vectors, respectively.
    \item \( \|V\|_2 \) and \( \|T\|_2 \) denote their \( L_2 \)-norms (Euclidean norms).
    \item \( \cdot \) represents the dot product between the vectors.
\end{itemize}

The cosine similarity score ranges from \( -1 \) (completely dissimilar) to \( +1 \) (identical), with higher values indicating stronger text-video alignment. Normalizing the vectors ensures that the similarity measure depends only on their angular distance rather than magnitude.

\noindent\textbf{CameraCLIP Loss}.  
To further align video and text modalities, we employ the InfoNCE loss, which maximizes the similarity of matching pairs while minimizing it for non-matching pairs (see Supplementary Material for details).

\subsection{CameraDiff and CLIPLoRA}
\label{sec:cameradiff}

As illustrated in Figure~\ref{fig:pipeline} (c), CameraDiff first takes classified videos from the 20 categories in Cinematic2K as input, using LoRA to train the T2V model on categorized cinematic pattern videos, enabling single-shot control. Each LoRA module encapsulates a distinct cinematic style, ranging from fundamental shot framing and angles to complex movements such as long shot, dutch angle and rack focus. More generated results are presented in Figure~\ref{fig:single_lora}.

Furthermore, we aim to enable multi-camera movement control within a single video. While previous multi-LoRA composition methods, including LoRA Merge, Switch, and Composite, have made notable progress in image generation, they rely on static, manually designed strategies. These approaches struggle to model the dynamic interactions between multiple LoRAs, leading to spatiotemporal inconsistencies and visual artifacts in video generation. To overcome these limitations, we propose CLIPLoRA, a method specifically designed for cinematic video generation that enables adaptive multi-LoRA composition.

We aim to determine the optimal LoRA to activate at each diffusion step during the denoising process. Given \( k \) available LoRAs and \( H \) diffusion steps, the exhaustive search space of \( k^H \) permutations is computationally infeasible. Inspired by LoRA Switch and genetic algorithms~\cite{goldberg1989genetic,alam2020genetic}, we represent each individual in the population as a sequence of LoRA compositions activated across the diffusion steps, sampled from the set \( P \) of available LoRAs. This adaptive approach efficiently explores the optimal LoRA activation sequence, balancing computational complexity and generation quality. The algorithm proceeds as follows:
\begin{enumerate}
    \item \textbf{Initialization:} Randomly sample an initial population of \( N \) LoRA composition sequences \( L_i \), \( i \in \{1, 2, \dots, N\} \), each representing a LoRA assignment across all diffusion steps.
    \item \textbf{Evaluation:} Generate videos based on a set of test prompts \( S \) using each sequence \( L_i \), and compute the average CameraCLIP score as the fitness:
    \begin{equation}
        F(L_i) = \frac{1}{|S|} \sum_{s \in S} \text{CameraCLIP}(L_i, s).
    \end{equation}
    \item \textbf{Selection:} Select the top \( p\% \) of individuals with the highest fitness scores as elite candidates for the next generation.
    \item \textbf{Crossover:} Perform crossover among selected individuals, exchanging segments of LoRA sequences between pairs of parents to generate diverse offspring.
    \item \textbf{Mutation:} With mutation probability \( p_m \), randomly alter individual genes (LoRA assignments at specific diffusion steps) to further explore the search space.
    \item \textbf{Iteration:} Repeat steps 2–5 for \( T \) generations until convergence or a predefined stopping criterion is satisfied.
\end{enumerate}

This approach enables an efficient search for optimal LoRA configurations, ensuring high-quality video generation while significantly reducing the computational cost.

\section{Experiment}
In this section, we evaluate the performance of CameraCLIP and CameraDiff. We begin by outlining the basic experimental settings. Next, we assess CameraCLIP's video-to-text retrieval accuracy on the validation dataset derived from Cinematic2K (see Section~\ref{sec:cameraclip-e}). We then present the qualitative and quantitative results of CameraDiff, including single-shot generation and multi-camera motion composition using CLIPLoRA (see Section~\ref{sec:cliplora-e}). Finally, we conduct an ablation study to analyze our temporal modeling method (see Section~\ref{sec:ablation:tmp}) and CameraCLIP's performance as the evaluator in CLIPLoRA (see Section~\ref{sec:ablation:cameraclip}).
\subsection{Setup}
\textbf{Video CLIP Models.} We evaluate four representative models for video-text alignment: CLIP4CLIP~\cite{luo2021clip4clip}, ViCLIP~\cite{wang2023internvid}, LongCLIP~\cite{zhang2024longclip}, and VideoCLIP-XL~\cite{wang2024videoclipxl}. To assess their performance, we split Cinematic2K into a training set for fine-tuning CameraCLIP and a validation set for benchmarking various VideoCLIP models. Detailed dataset information is provided in the Supplementary Materials.

\noindent\textbf{Evaluation Metrics.} We evaluate the alignment between text and video using recall at 1 (R@1), which represents the proportion of queries for which the correct answer is found within the top-1 returned result. The quality of generated videos is evaluated using Fréchet Video Distance (FVD)~\cite{unterthiner2018towards_fvd} and CLIPSIM~\cite{Radford2021clip_icml}, which assess visual fidelity, temporal coherence, and alignment with cinematic language, respectively. FVD is computed using reference videos from WebVid10M~\cite{bain2021frozen}.

\noindent\textbf{LoRA Composition Baselines.} We compare our CLIPLoRA method with direct generation method (Origin) and several LoRA composition baselines: LoRA Merge, Switch, and Composite, for generating multi-shot control in a single video, using Animatediff~\cite{guo2023animatediff} as the backbone model. The detailed experimental settings for CameraDiff and the genetic algorithm used to search for optimal LoRA composition are provided in the Supplementary Material.

\subsection{CameraCLIP}
\label{sec:cameraclip-e}
In this section, we evaluate the performance of our proposed CameraCLIP on the R@1 metric through experiments with different representative video CLIP models and resolutions.



\begin{table}[t]
\centering
\renewcommand{\arraystretch}{1.2}
\vskip 0.15in
\small 
\begin{tabular}{c c c}
\toprule
\textbf{Models} & \textbf{Different Types} & \textbf{R@1 $\uparrow$} \\ 
\midrule
\multirow{4}{*}{CLIP~\cite{Radford2021clip_icml}} 
& ViT-B/32-224px & 0.63 \\ 
& ViT-B/16-224px & 0.64\\ 
& ViT-L/14-224px & 0.68\\ 
& ViT-L/14-336px & 0.71 \\
\midrule
\multirow{6}{*}{CLIP4CLIP~\cite{luo2021clip4clip}} 
& ViT-B/32 meanP & 0.71 \\
& ViT-B/32 seqLSTM & 0.69 \\
& ViT-B/32 seqTransf & 0.66 \\
& ViT-B/16 meanP & 0.72 \\
& ViT-B/16 seqLSTM & 0.70 \\
& ViT-B/16 seqTransf & 0.72 \\
\midrule
ViCLIP~\cite{wang2023internvid} & ViCLIP-B-16 & 0.64 \\
& ViCLIP-L-14 & 0.68 \\
\midrule
\multirow{2}{*}{LongCLIP~\cite{zhang2024longclip}} 
& LongCLIP-B & 0.55 \\
& LongCLIP-L & 0.75 \\
\midrule
\multirow{2}{*}{VideoCLIP-XL~\cite{wang2024videoclipxl}} 
& VideoCLIP-XL & 0.77 \\
& VideoCLIP-XL-V2 & 0.76 \\
\midrule
\multirow{4}{*}{CameraCLIP (Ours)} 
& ViT-B/32-224px & 0.69 \\
& ViT-B/16-224px &  0.73\\
& ViT-L/14-224px & 0.77\\
& ViT-L/14-336px & \textbf{0.83} \\
\bottomrule
\end{tabular}
\caption{\textbf{Quantitative comparison with other video CLIP models}. CameraCLIP outperforms other models on the validation set derived from the Cinematic2K, achieving the best performance. The best-performing result is highlighted in bold.}
\label{tab:clip_baseline}
\vspace{-2em}
\end{table}




As shown in Table~\ref{tab:clip_baseline}, traditional CLIP models with varying Vision Transformer (ViT) sizes and input resolutions exhibit moderate accuracy, with larger models and higher resolutions yielding better performance (e.g., ViT-B/32-224px achieves 0.63, while ViT-L/14-336px reaches 0.71). CLIP4CLIP and ViCLIP, which incorporate additional layers or sequence models, offer slight improvements, peaking at 0.72. LongCLIP benefits from extended input length, achieving an R@1 score of 0.75, while VideoCLIP-XL further improves to 0.77 using a text-similarity-guided primary component matching mechanism.

Our proposed CameraCLIP surpasses all baselines, achieving an R@1 score of 0.83 with the ViT-L/14-336px configuration, demonstrating superior understanding of cinematic language, including shot framing, shot angles, and camera movement. Furthermore, CameraCLIP consistently outperforms traditional CLIP models across all configurations, confirming that fine-tuning on a specialized cinematic dataset enhances video-specific linguistic alignment. These results validate CameraCLIP’s effectiveness in bridging cinematic language with video content.

\subsection{CameraDiff}
\label{sec:cliplora-e}

The qualitative results of CameraDiff for single-shot generation are presented in Figure~\ref{fig:single_lora}. After obtaining LoRA weights for specific cinematic styles, we apply CLIPLoRA to compose multiple shots within a single video. The corresponding qualitative results are shown in Figure~\ref{fig:multi_lora}.
\begin{figure}[t]
    \centering
    \includegraphics[width=\columnwidth]{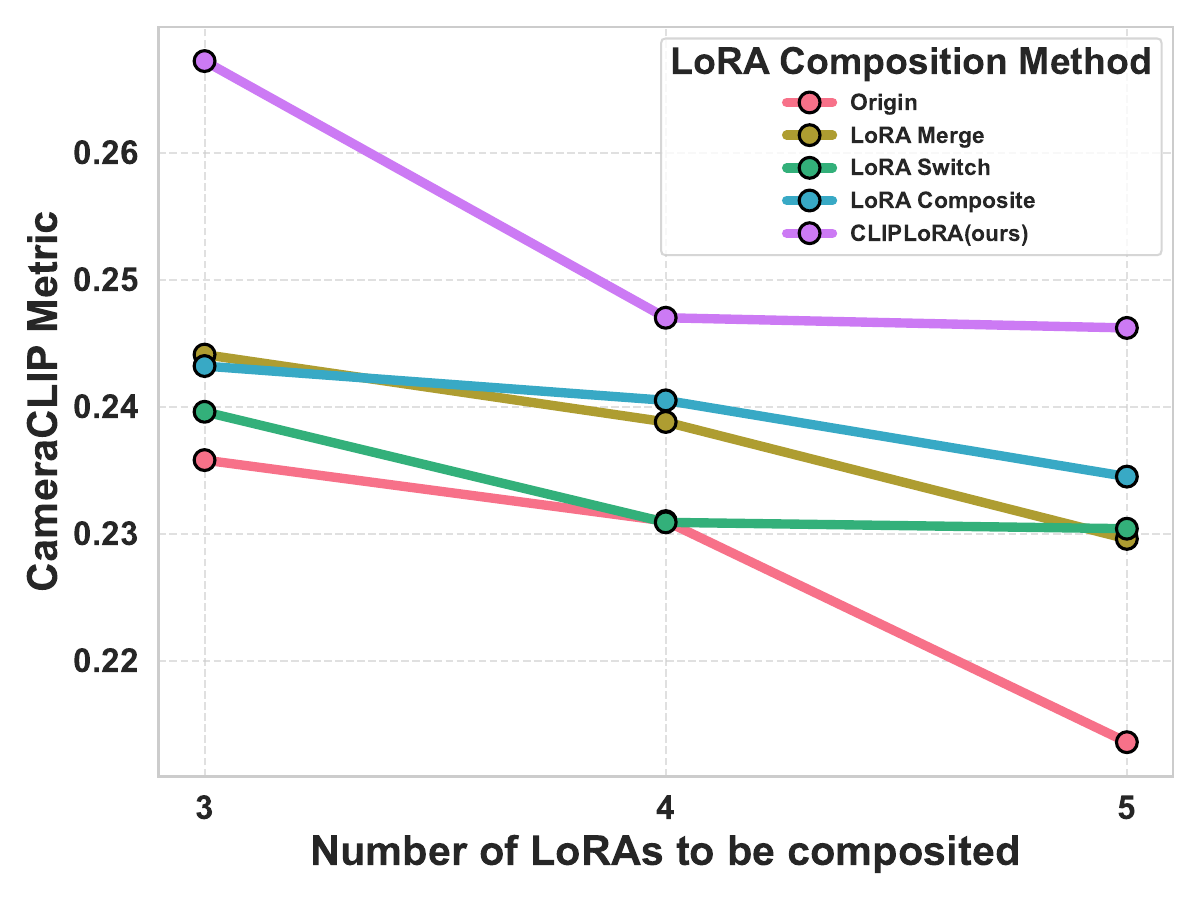}
    \caption{Comparison of CLIPLoRA results with other LoRA composition methods.}
    \vspace{-1.8em}
    \label{fig:lora_composition}
\end{figure}

First, we use CLIPSIM to measure cinematic text-video consistency as the number of LoRAs increases, as shown in Figure~\ref{fig:lora_composition}. CLIPLoRA consistently outperforms all baselines while maintaining stable performance, whereas traditional static methods suffer from detail loss and disjointed transitions, particularly with a higher number of LoRAs. These results highlight the effectiveness of CLIPLoRA for multi-shot integration.

Additionally, we use FVD to evaluate the generated video's visual fidelity and temporal coherence, as shown in Table~\ref{tab:fvd_clip_baseline}. The results demonstrate that CLIPLoRA significantly outperforms existing methods in both FVD and CLIPSIM, confirming its superiority in generating high-quality composite multi-shot videos.

\begin{table}[htbp]
\centering
\small 
\renewcommand{\arraystretch}{1.4} 
\setlength{\tabcolsep}{6pt} 
\resizebox{\linewidth}{!}{
\tablestyle{3pt}{1.4}
\begin{tabular}{lccccc}
        \toprule
        & \textbf{Origin} & \textbf{Merge} & \textbf{Switch} & \textbf{Composite} & \textbf{CLIPLoRA} \\
        \midrule
        \textbf{FVD$\downarrow$} & 2534 & 2358 & 2730 & 2404 & \textbf{1837} \\
        \textbf{CLIPSIM$\uparrow$}  & 0.2268 & 0.2375 & 0.2336 & 0.2394 & \textbf{0.2535} \\
        \bottomrule
\end{tabular}
}
\caption{\textbf{Quantitative comparison with other LoRA composition methods.} Our proposed CLIPLoRA surpasses existing approaches in both text-video similarity and generated quality of cinematic language videos, with the highest score highlighted in \textbf{bold}.}
\label{tab:fvd_clip_baseline}
\vspace{-4mm}
\end{table}

\subsection{Ablation Study}
\subsubsection{Temporal Modeling Method}
\label{sec:ablation:tmp}

To validate the effectiveness of the mean pooling approach for modeling video frames, we conducted an ablation study comparing various temporal modeling methods, including Transformer, LSTM, MLP, Multi-Head Attention, and Transformer+LSTM. As shown in Table~\ref{tab:clip_ablation}, Transformer achieved the second-best R@1 score of 0.43, while mean pooling significantly outperformed it with an accuracy of 0.81. This suggests that, for our small dataset task, mean pooling provides a more robust frame representation, as complex models tend to overfit and reduce generalization, consistent with prior work's conclusion~\cite{luo2021clip4clip}.



\begin{table}[ht]
    \centering
    \begin{tabular}{cc} 
        \toprule
        \textbf{Temporal Modeling Methods} & \textbf{R@1 $\uparrow$} \\
        \midrule
        Transformer & 0.43 \\
        LSTM & 0.34 \\
        MLP & 0.29 \\
        Multi-Head Attention & 0.39 \\
        Transformer+LSTM & 0.32 \\
        \midrule
        Mean Pooling(Ours) & \textbf{0.83} \\
        \bottomrule
    \end{tabular}
    \vspace{-0.5em}
    \caption{\textbf{Quantitative comparison with other temporal modeling methods to train CameraCLIP}. We evaluate their performance on the cinematic validation set using the R@1 score, with the highest accuracy highlighted in \textbf{bold}.}
    \label{tab:clip_ablation}
\end{table}

\subsubsection{CameraCLIP as Evaluator}
\label{sec:ablation:cameraclip}

\begin{table}[ht]
\centering
\small 
\renewcommand{\arraystretch}{1.4} 
\setlength{\tabcolsep}{6pt} 
\resizebox{\linewidth}{!}{
\tablestyle{3pt}{1.4}
 \begin{tabular}{lcccc}
        \toprule
        & \textbf{CLIP} & \textbf{LongCLIP} & \textbf{VideoCLIP-XL} & \textbf{CLIPLoRA} \\
        \midrule
        \textbf{FVD$\downarrow$} & 1958 & 1957 & 1921 & \textbf{1837} \\
        \textbf{CLIPSIM$\uparrow$} & 0.2396 & 0.2417 & 0.2427 & \textbf{0.2535} \\
        \bottomrule

\end{tabular}
}
\caption{\textbf{Quantitative comparison with other video CLIP models.} Our proposed CameraCLIP serves as an evaluator to guide multi-LoRA selection during the denoising stage, outperforming other video CLIP models in both text-video similarity and cinematic video quality, with the highest score highlighted in \textbf{bold}.}
\label{tab:fvd_ablation}
\vspace{-2em}
\end{table}

Table~\ref{tab:fvd_ablation} presents a quantitative comparison of our proposed CameraCLIP with other video CLIP models used as evaluators in the search for optimal LoRA compositions. CameraCLIP outperforms the other models in both CLIPSIM and FVD, demonstrating that, by guiding multi-LoRA selection during the denoising stage, CameraCLIP effectively enhances both text-video similarity and cinematic video generation quality. Furthermore, as shown in Table~\ref{tab:fvd_clip_baseline}, the dynamic LoRA composition methods outperform the traditional static composition approach, suggesting that, in video generation, a carefully crafted LoRA merge strategy can significantly improve video quality. This also provides insight for researchers focusing on the LoRA composition stage to enhance customized video quality.
\section{Conclusion}
In this work, we focus on enhancing the capability of T2V models to generate cinematic language videos. To achieve this, we propose a threefold solution. First, we introduce Cinematic2K, a comprehensive dataset encompassing shot framing, shot angles, and camera movements. Second, we present CameraDiff, which leverages LoRA for precise cinematic control. Finally, to improve multi-shot composition, we propose CLIPLoRA, using CameraCLIP as the evaluator to guide dynamic LoRA composition, facilitating smooth transitions and realistic blending of cinematic styles within a single video.

Experimental results show that CameraCLIP achieves an R@1 score of 0.83, surpassing existing models in cinematic text-video alignment. CameraDiff ensures stable single-shot control, while CLIPLoRA integrates multiple LoRAs to generate complex cinematic compositions and provides insights for enhancing customized video quality. These contributions establish a foundation for controllable cinematic language generation, bridging the gap between automated video synthesis and expert cinematography.
{
    \small
    \bibliographystyle{ieeenat_fullname}
    \bibliography{main}
}


\end{document}